\pdfoutput=1
\documentclass[11pt]{article}

\usepackage{acl}
\usepackage{times}
\usepackage{latexsym}

\usepackage[T1]{fontenc}
\usepackage[utf8]{inputenc}

\usepackage{microtype}
\usepackage{caption}
\usepackage{subcaption}
\usepackage{graphicx}
\usepackage{multirow}
\usepackage{amsmath, nccmath}
\usepackage{arydshln}
\usepackage{enumitem}
\usepackage[normalem]{ulem}

\title{Learning to Mediate Disparities Towards Pragmatic Communication}

\author{Yuwei Bao$^\dagger$ \hspace{30pt} Sayan Ghosh$^\mathsection$\thanks{\ \ Work done during undergraduate study at the University of Michigan.}  \hspace{30pt}  Joyce Chai$^\dagger$\\
$^\dagger$Computer Science and Engineering, University of Michigan \hspace{5pt}
$^\mathsection$ASAPP \\
  \texttt{\{yuweibao, sayghosh, chaijy\}@umich.edu}  
}

\begin{document}
\maketitle
\begin{abstract}
Human communication is a collaborative process. Speakers, on top of conveying their own intent, adjust the content and language expressions by taking the listeners into account, including their knowledge background, personalities, and physical capabilities. Towards building AI agents with similar abilities in language communication, we propose Pragmatic Rational Speaker (PRS), a framework extending Rational Speech Act (RSA). The PRS attempts to learn the speaker-listener disparity and adjust the speech accordingly, by adding a light-weighted disparity adjustment layer into {\em working memory} on top of speaker’s {\em long-term} memory system. By fixing the long-term memory, the PRS only needs to update its working memory to learn and adapt to different types of listeners. To validate our framework, we create a dataset that simulates different types of speaker-listener disparities in the context of referential games. Our empirical results demonstrate that the PRS is able to shift its output towards the language that listeners are able to understand, significantly improve the collaborative task outcome. 
%and learn the disparity more efficiently than joint training.
\end{abstract}

\section{Introduction}

%\jycc{I think the introduction needs to be re-structured. I'll come back to it later. Have to go to a meeting}. 

In human communication, speakers often adjust their language production by taking into consideration listeners' personality, background knowledge, perceptual or physical capabilities etc~\cite{clark_1996}. 
%The speaker makes this extra effort so that the listener would understanding what's been communicated~\cite{clark96}. 
%Their understanding of each other's knowledge and ability shape how they communicate with each other in order to reach common goals~\cite{clark96}. \jycc{add citation on herbert clark, using language, check paul's paper}
%
Recent years have seen an increasing amount of work that explores pragmatic reasoning based on Rational Speech Act (RSA) \cite{andreas, fried2018unified, friedspeaker, whitelearning, cohnpragmatically},  multi-agent emergent communication framework~\cite{lazaridou, lazaridou2}, and Theory of Mind in communication~\cite{paul,zhu2021few}. However, except for \cite{zhu2021few}, most previous works assume that the listeners and the speakers have the same background knowledge and capabilities, including vocabulary size, visual access, and relative locations. This assumption is a great simplification of real-world communication where speakers and listeners often have various types of disparities.

\begin{figure}
     \centering
     \begin{subfigure}[b]{0.23\textwidth}
         \centering
         \includegraphics[width=0.98\textwidth]{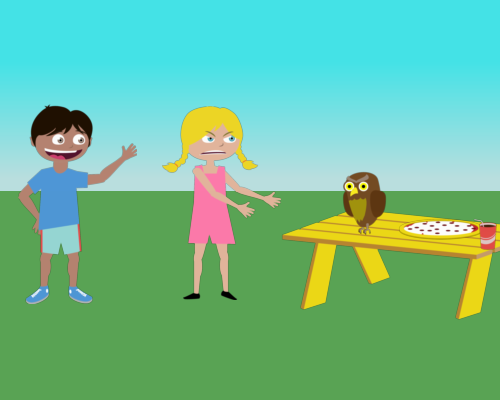}
         \caption{Target}
         \label{1t}
     \end{subfigure}
     \hfill
     \begin{subfigure}[b]{0.23\textwidth}
         \centering
         \includegraphics[width=0.98\textwidth]{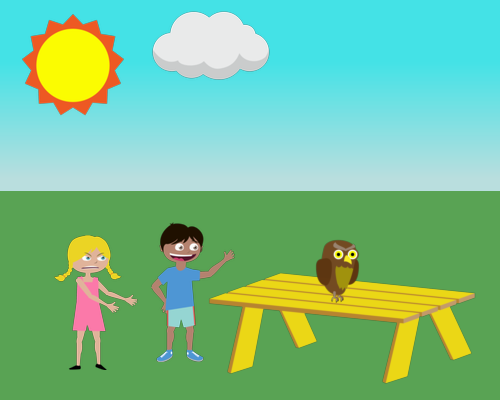}
         \caption{Distractor}
         \label{1d}
     \end{subfigure}
     \hfill 
     \begin{subfigure}[b]{0.5\textwidth}
         \small{.\\\\
         \textbf{Literal Speaker}: There is an owl on the table.\\
         \textbf{Rational Speaker}: There is a \underline{pizza} on the table.\\
         \textbf{Listener's Disparity}: understands hypernym of food only.\\
         \textbf{Pragmatic Rational Speaker}: There is \underline{food} on the table.}
     \end{subfigure}
    \caption{\small \textbf{TASK:} \underline{Given two images, the speaker generates} \underline{a description for the target image and asks the listener to pick} \underline{out the image described. Both players win if the listener} \underline{picks the correct one.} In this example, a Literal Speaker could generate multiple captions that suit the target, such as the one above, whereas a Rational Speaker limits the description to the unique features of the target (e.g. pizza). If the listener only understands the hypernym of food (disparity), a Pragmatic Rational Speaker would learn the disparity and use {\em food} instead of {\em pizza} to help the listener understand. }
   %\vspace{-20pt}
     \label{imgexp}
\end{figure}

%The disparities between the speaker and the listener across these factors created a gap in reaching the common communicative goal, and takes extra steps to bridge the gap. Each conversation is a modification on the long-term belief \cite{longm} with situation-specific factors.  

%However, most prior work with Rational Speech Act (RSA) based neural listener/speaker pragmatic reasoning \cite{andreas, fried2018unified, friedspeaker, whitelearning, cohnpragmatically} and natural language multi-agent communication work \cite{lazaridou, lazaridou2} assumes that the listeners and the speakers have the exact same background knowledge and capabilities, including vocabulary size, visual access, and relative locations (this is not a limitation of the RSA model itself, however). This assumption is a great simplification of real-world communication where speakers and listeners often have various types of disparities. 

To address this limitation, this paper extends the Rational Speech Act (RSA)~\cite{rsa} model towards rational agents learning to adapt behaviors based on their experience with the listener. 
The design choice of our model is inspired by the human cognitive system \cite{workingm, Wardlow} where a limited capacity \textbf{\textit{working memory}} is built on top of the \textbf{\textit{long-term memory}} to adjust the output to be task and environment specific. Each communication is a modification on the long-term memory \cite{longm} with situation-specific factors. In our framework, we fix the long-term memory which captures language structure for communication, and introduce a light-weighted working memory \cite{workingm2} for the Pragmatic Rational Speaker to modify and accommodate two goals: 1) a \textbf{ task goal} which retrieves relevant information from the long-term memory and accomplish the task, and 2) a \textbf{ disparity goal} which learns and adjusts the conversation to accommodate the listener's disparity through reinforcement learning. We separate each component as they are independent of each other in utility, and can be easily switched and adapted for new tasks and new environment.

Different from previous works which only demonstrate how learned models affect task performance (e.g. \cite{alfworld, zhu2021few, Corona}), one of our goals is to also provide transparency on what models have indeed learned towards the end goal. It’s well established that end-to-end neural models can often take advantage of spurious data bias to gain end performance. Models that only report end measure without showing their internal works would not be sufficient to tell the whole story about model’s abilities. 

To serve this goal, we situated our investigation in the context of a referential game\footnote{Different from traditional referential ground work as \cite{refgame1,refgame3,refgame4,refgame6,refgame7}, we adopted this term from a recent line of work \cite{lazaridou,andreas} to refer to the task described in Figure \ref{imgexp}.} 
as shown in Figure~\ref{imgexp}. We carefully curated a dataset to simulate two types of disparity: \textit{knowledge disparity} and \textit{perceptual disparity}. 
Our empirical results demonstrate that our model is able to significantly improve the collaborative game performance by shifting communication towards the language that the listeners with disparities are able to understand. In addition, our results show that separating working memory from long-term memory leads to faster learning and better performance than the previous model which conducted joint end-to-end learning.
% The efficiency comparison also shows the light weighted working memory can learn much faster and better than joint training.

Our contributions are the following. 1) Following human cognition, we demonstrate the benefits of separating working memory from the long-term memory, compared to end-to-end joint training. 2) We propose a new dataset to simulate multiple distinct types of disparities, and demonstrate the pragmatic adaptability of our model. 3) Instead of focusing on mere end task performance, we show model's strong language shift ability to accommodate listener's disparities. The dataset and code are available through \url{https://github.com/sled-group/Pragmatic-Rational-Speaker} to facilitate future work on pragmatics and theory of mind in language interpretation and generation.

% 3) To the best of our knowledge, we are among the first to demonstrate model's strong language shift capability to actually accommodate listener's disparities instead of a simple end task performance measure.

%\vspace{-10pt}
%\begin{itemize}
%    \item A pragmatic multi-agent communication model to accommodate agent disparities
%   \vspace{-10pt}
%    \item A novel application of the working memory for efficient situated language adjustment and pragmatic inference
 %\vspace{-10pt}
%\end{itemize}

\section{Related Work}

It has been studied \cite{Leung,Stephens,Wardlow} in psychology that human speakers adjust the way how we speak for successful communication after learning the listener's disparity. Some recent work \cite{similar,zhu2021few,Corona,Hawkins} attempt to address similar questions. We build our model upon the following two concepts.

\subsection*{Rational Speech Act (RSA)}

The Rational Speech Act (RSA) model \cite{rsa} is a probabilistic model for the speakers and listeners to pragmatically reason about each other's intention. In the context of a referential game \cite{rsa2}, for example (Figure \ref{imgexp}), given an image $m$, it starts with a literal speaker $S_0$ to generate caption $c$: $P_{S_0}(c|m)$. A rational listener $L_1$ reasons about the literal speaker's ($S_0$) strategy and picks the best image that matches the description. A rational speaker $S_1$ then takes the rational listener's ($L_1$) strategy into account and produces a caption $c$ that maximizes the collaborative game goal. 

%\vspace{-10pt}
$$P_{L_1}(m|c) \propto P_{S_0}(c|m) \cdot P(m)$$
$$P_{S_1}(c|m) \propto P_{L_1}(m|c) \cdot P(c)$$
%\vspace{-15pt}

In previous work \cite{andreas} and \cite{lazaridou, lazaridou2}, the same referential game setup was used to propose a rational speaker that learns to reason the collaborative game and to produce natural sounding image captions based on RSA. However, they were mainly addressing the \textit{task goal}, assuming the speaker and listener have the exact same capabilities and knowledge background, which is unrealistic. In our work, we created listeners with disparity $d$ and extend this model for the speaker to accommodate both the \textit{task} and \textit{disparities} goals.

% \ybed{Different from previous work (e.g. Chai 13/14) which resolve referring expressions to individual objects, recent work (cited above) also use it to describe individual images among distractors. In our dataset, multiple objects are available in each image for the speaker to choose from, and we challenge the speaker to pick the suitable ones to distinguish the target image and for the listeners to understand.}

% \subsection*{Emergent Communication}
%  Our work also draws from recent work in emergent communication \cite{lazaridou, lazaridou2, Foersteretal2016, shah-etal-2018-bootstrapping, kottur-etal-2017-natural, similar, zhu2021few, Corona}. These works involve reinforcement-based learning to reach a common goal in a collaborative game. Our work is inspired by these for the Pragmatic Rational Speaker to realistically learn the listener's disparities through communication interactions, without any prior knowledge on the listener's background nor an oracle access to probe the listener's brain. 

\begin{table*}[]
    \centering
    \begin{tabular}{|c|l|lll|}
    \hline
    & Role & Long Term & Work Mem - Task & Work Mem - Disparity\\\hline
      $S_0$  & Literal Speaker   &  Image Caption & &\\
      $S_1$  & Rational Speaker   & Image Caption & Simulated Listener & \\
      $S_1^d$ & Pragmatic Rational Speaker    &  Image Caption & Simulated Listener & Disparity Adjustment\\ \hline
    %   & Rational Speaker No Disparity    &  Rational Speaker ($S_1$) + Same Disparity as Corresponding Listener\\
      $L_1$ & Rational Listener    &  \multicolumn{3}{c|}{Caption Grounding}\\
      $L_1^d$ & Rational Listener w/ Disparity   & 
      \multicolumn{3}{c|}{Caption Grounding with disparity}
      \\\hline
    \end{tabular}
    \caption{Types of Speaker and Listener}
    \vspace{-8pt}
    \label{sl}
\end{table*}

\subsection*{Working Memory}

Working memory (also short-term memory) is used in neuropsychology and cognitive science \cite{workingm, workingm2} to refer to the memory that controls attention, plans and carries out behavior. It is a combination of multiple components, including the contribution of long-term memory \cite{longm, longmC} and situation-specific task processing \cite{workingmC}. 

The classical artificial intelligence work such as ACT \cite{act} and SOAR \cite{soar} also incorporated the concept of working memory to model human short-term memory. The similar concept has been used in recent work such as \cite{felix1, felix2}. Our work is a novel application of the working memory to pragmatically adjust communication for speaker-listener disparities (\textit{disparity} goal), and take advantage of the internal simulation architecture to achieve the \textit{task} goal.

% \subsection*{Theory of Mind}
% It has been studied \cite{Leung,Stephens,Wardlow} in psychology that human speakers adjust the way how we speak for successful communication after learning the listener's disparity. Some recent work \cite{similar,zhu2021few,Corona,Hawkins} attempt to address similar questions. 

\hfill

Similar to \cite{kottur-etal-2017-natural, lazaridou}, our model learns to converge language to adapt to listener's disparities through interactions, instead of ground truth supervision on language generation. The speakers have zero prior knowledge on the listener's background nor an oracle access to probe the listener's brain.

Different from previous works, 
%some of which used meta-learning to study the disparity within the same family (e.g. different lexical levels of multi-linguistic learning), 
our model is able to generalize to distinct types of disparities. In addition, while previous models were trained in an end-to-end joint fashion, our work separates training and demonstrates the efficiency of working memory. Most importantly, few of the previous work were able to showcase model's language capabilities and only evaluate them by the end performance (e.g. accuracy), whereas our work emphasizes on evaluating how well the models learn to shift the language towards better understanding.

% A few recent works \cite{zhu2021few}, [TODO: cite Corona et al] proposed meta-learning on a single family of disparity adjustments (e.g. different lexical levels of multi-linguistic learning) through interactions. Different from that, we propose a more generalized framework, that can be used to adjust for distinct types of disparities across various tasks. Moreover, the previous works rely on end-to-end joint training that are inflexible and inefficient for new tasks nor disparities. Different from previous works, we emphasize on the strong communication abilities and demonstrate models' actual language adjustments on top of simple accuracy improvement.

% One example of working memory is the Theory of Mind (ToM) \cite{tom}, where the speaker makes predictions on the listener's mind based on observations and interactions. \cite{zhu2021few}'s work used this concept to model listener's mental state, in order to achieve the collaborative game goal. Different from that, our work separates the task specific working memory from listener's disparity adjustment ToM to achieve lighter weighted adjustment, and uses state-of-the-art image captioning components to demonstrate the actual language distribution shift (i.e. the concrete vocabulary adjustment) on top of conventional accuracy performance.

\section{Dataset}\label{dataset}

There are many levels of disparities during verbal communication \cite{Stephens}, including phonetic, lexical, grammatical, semantic representations, etc. In our work, we assembled two datasets, and challenge the speaker model to handle two types of disparities: 1) \underline{knowledge disparity}, and 2) \underline{perceptual disparity}. 

The {\em knowledge disparity} is simulated through the \textit{hypernym} dataset, where the listener only understands the hypernym for all the objects (e.g. ``food'' instead of ``pizza''), whereas the speaker understands both. This dataset challenges the speaker model at the \textit{lexical} level to learn what listener's vocab limitation, and shift towards the words that they understand. 

The {\em perceptual disparity} is simulated through the \textit{limited visual} dataset, where the listener has impaired vision or some objects were physically blocked from the eyesight. This dataset challenges the speaker
to shift attention and pick the visible objects for the listener to describe. 
For control and demonstration purposes, we remove all the animal-related objects and words from listener's training. 

These datasets are used to simulate listener's disparities and train the listener's model as described in Section \ref{l1}. The speaker's long term memory was trained with the original data which has full knowledge of the vocab and objects, but no idea what the listeners are or aren't capable of. Detailed dataset components can be found in the Appendix.

We modified the Abstract Scenes \cite{Ortiz} dataset for our experiments. There are 10020 images, each including 3 ground truth captions, and a median of 6 to 7 objects. We assembled $\sim$35k pairs of images that differ by $\leq4$ objects as the \textit{Hard} set, $\sim$25k pairs that differ by $>4$ objects as the \textit{Easy} set, and together as the \textit{Combined} set. The image pairs were split into training, validation and testing by a ratio of 8:1:1.  

% \begin{table}
% \small
% \centering
%   \begin{tabular}{|l|l|l|l|}
%     \hline
%   \textbf{Disparity} & \textbf{Mod Modality} & \textbf{Speaker} & \textbf{Listener}\\
%     \hline \hline
%     Knowledge & Text & Original & Hypernym \\ \hline
%     Visual & Image, Text& Original & Original-Animal\\ \hline
%   \end{tabular}
%       \caption{Disparity Simulation Datasets}
%     \label{data}
% \end{table}

\section{Method}

\begin{figure*}[t]
     \centering
     \includegraphics[width=\textwidth]{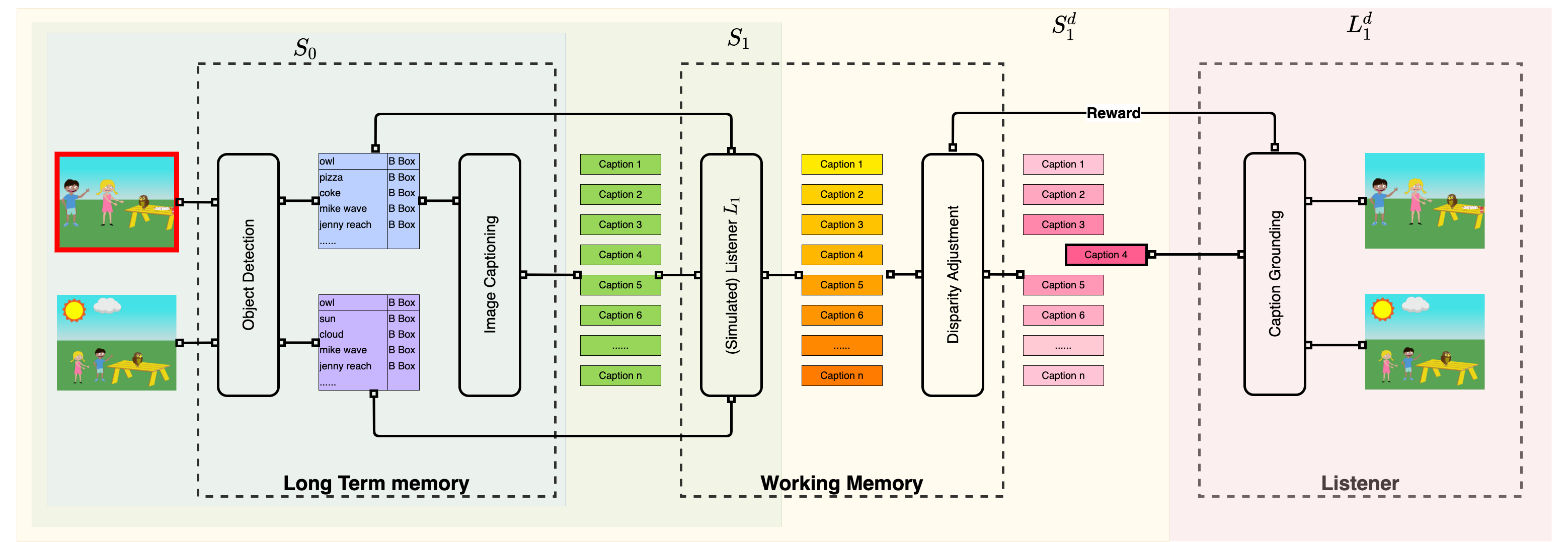}
     \caption{Speaker and Listener Models: Literal Speaker $S_0$ uses and object detector and image captioning module to generate a list of candidate captions in the fixed long term memory. The Rational Speaker $S_1$ simulates an internal listener to rank (illustrated by color gradient) the candidate captions by their uniqueness in describing the target image. The Pragmatic Rational Speaker $S_1^d$ interacts with the actual listener to rerank the captions and pick out the best one to accommodate the disparity and the task goal. Both simulated listener and disparity adjustment components are inside the working memory for task specific and disparity specific adjustments.}
     %\vspace{-15pt}
     \label{smodel}
\end{figure*}

Given a pair of images $m_0, m_1$, the target image indicator $t\in\{0, 1\}$, and the listener's disparity $d$, the speaker generates a caption $c$ for the target image $m_t$, and the listener needs to pick out the correct target $t$ given $c$. Both receive a reward of $+1$ upon correct choice, and $-1$ otherwise.

%\ybed{ (Figure \ref{smodel}, Table \ref{sl}), 
Following the RSA model, as shown Figure \ref{smodel}, we start by building the Literal Speaker $\mathbf{S_0}$, gradually increase model structure and functionality with the vanilla Rational Speaker $\mathbf{S_1}$ and the Pragmatic Rational Speaker $\mathbf{S_1^d}$. Upon retrieving a list of candidate captions $C$ from the long-term memory, the final goal for $\mathbf{S_1^d}$ is to output the best caption $c$ in the working memory, that accommodates both 1) \textbf{task goal:} describes the unique features of the target image, and 2) \textbf{disparity goal:} learns and accommodates the listener's disparity.

Table \ref{sl} is a brief summary of each model. The Literal Speaker $\mathbf{S_0}$ generates candidate captions $c$ for a given image $m$ (Eq \ref{eq_s0}), which serves as the long-term memory. The Rational Listener $\mathbf{L_1}$ picks out an image as the target given speaker's description (Eq \ref{eq_l1}). The vanilla Rational Speaker $\mathbf{S_1}$ achieves the \textit{task goal} by simulating the listener's mind internally in its working memory (Eq \ref{eq_s1}). $\mathbf{L_1^d}$ incorporates disparity to the Rational Listener. The Pragmatic Rational Speaker $\mathbf{S_1^d}$ adds a light-weight disparity adjustment layer (Eq \ref{eq_s1d}) to learn and accommodate listener's disparity through interactions, and achieves both goals. Each component can be easily switched and adapted to new tasks or environment. 

\begin{flalign}\label{eq_s0}
    \mathbf{S_{0}}: P(c | m_t) &&
\end{flalign} 
% \vspace{-20pt}

\begin{flalign}\label{eq_l1}
    \mathbf{L_1}: P(t|m_0, m_1, c) \propto P_{S_0}(c|m_t) \cdot P(m_t) &&
\end{flalign}
% \vspace{-20pt}

\begin{fleqn}
\begin{equation}\label{eq_s1}
\begin{split}
    \mathbf{S_1}: &P(c|m_0, m_1, t) \propto\\
    &P_{L_1}(t|m_0, m_1, c) \cdot P(c|m_0, m_1)
\end{split}
\end{equation}
\end{fleqn}
% \vspace{-20pt}

\begin{fleqn}
\begin{equation}\label{eq_l1d}
\begin{split}
    \mathbf{L_1^d}: &P(t|m_0, m_1, c, d)\propto\\
    &P_{S_1}(c|m_0, m_1, t, d) \cdot P(t| m_0, m_1, d)
\end{split}
\end{equation}
\end{fleqn}
% \vspace{-25pt}

\begin{fleqn}
\begin{equation}\label{eq_s1d}
\begin{split}
    \mathbf{S_1^d}: &P(c|m_0, m_1, t, d)\propto\\ 
     &P_{L_1^d}(t|m_0, m_1, c, d) \cdot P(c|m_0, m_1, d)
\end{split}
\end{equation}
\end{fleqn}

\subsection{Literal Speaker $S_0$}\label{s0}
The Literal Speaker $S_0$ (Figure \ref{smodel}) is an object detection based image captioning module that generates caption candidates for the target image.

% \begin{equation}
%     S_{0}: P(c | m) := \mbox{Transformer}(\mbox{ObjDec}(m))
% \end{equation}
%\vspace{-8pt}
\begin{equation}
\begin{split}
    &o_1,\dotsc, o_k, b_1,\dotsc, b_k = \mathtt{ObjDet}(m_t)\\
    &e_1,\dotsc, e_k = \mathtt{WordEmb}(o_1, \dotsc, o_k)\\
    &c_1,\dotsc, c_n = \mathtt{Transformer}(e_1,\dotsc, b_1,\dotsc)
\end{split}
\end{equation}
%\vspace{-10pt}

For a given target image $m_t$, since it's important to ground words to the scenes in order to control the disparities in vocabularies, we applied the object detector YOLO3 \cite{yolo3} to extract a list of $k$ detected objects $O=\{o_1, o_2, \dotsc, o_k\}$, and their corresponding bounding boxes $B=\{b_1, b_2, \dotsc, b_k\}$. Each image chooses at most $max\_obj=9$ detected objects, and the names of each were embedded with a pre-trained BERT \cite{bert} word embedding $E = \{e_1, e_2, \dotsc, e_k\}$. These embeddings are then concatenated with their bounding box locations, and sent to the Transformer Decoder to generate $beam\_size = 30$ candidate captions $C = \{c_1, c_2, \dotsc, c_n\}$ for each target image.

% Since it's important to ground words to the scenes in order to control the disparities in vocabularies, we applied the object detector YOLO3 \cite{yolo3} for image embedding. Each image chooses at most $max\_obj=9$ detected objects, and the names of each were embedded with a pre-trained BERT \cite{bert} word embedding. These embeddings are then concatenated with their bounding box locations to offer information on the relative positions of each object. The embedded images were then sent to the Transformer Decoder to generate $beam\_size = 30$ candidate captions for each target image.

\subsection{Rational Listener ($L_1$)} \label{l1}
Without disparity concerns, the Rational Listener picks out the image that they believe is the target.
% \begin{equation}
%     L_1: P(t|m_0, m_1, c) = argmin_{t\in{0,1}} \mbox{CosSim}(\mbox{Transformer}(m_t, c), c)
% \end{equation}

% \begin{equation}
% \begin{split}
%  L_1: P(t|m_0, m_1, c) := argmin_{t\in\{0,1\}}\\ \mbox{CosSim}(\mbox{Transformer}(m_t, c), c)
% \end{split}
% \end{equation}
%\vspace{-5pt}
\begin{equation}
\begin{split}
&g_0 = \mathtt{FT\_Transformer}(m_0, c)\\
&g_1 = \mathtt{FT\_Transformer}(m_1, c)\\
&t = \mathrm{argmax}_{i\in\{0,1\}}\mathtt{CosSim}(g_i, c)
\end{split}
\end{equation}
%\vspace{-10pt}

Recall that $S_0$ used a $\mathtt{Transformer}$ decoder to connect the image and its corresponding captions. We reuse the same \underline{Fixed} pre-trained \underline{Training-mode} Transformer module (named $\mathtt{FT\_Transformer}$) to decide which image does the caption ground better in. Adopting the idea of teacher-forcing language training, the output ($g_i$) of $\mathtt{FT\_Transformer}$ with an input pair ($m_i, c$) should closely resemble the original input $c$ if the input image $m_i$ is indeed the one used to generate the caption $c$. By calculating the cosine similarity of each ($g_i, c$) pair, the image that grounds better (higher $\mathtt{CosSim}$) in the description would be chosen as the target.

% We use a \underline{fixed} pre-trained \underline{training-mode} image caption model ($\mathtt{FT\_Transformer}$) used by $S_{0}$ to decide which image the caption grounds better in. Given an encoded input image (e.g. $m_0$) and a caption $c$, the fixed transformer decoder has been trained to ground the image captions to their corresponding images. Adopting the idea of teacher-forcing language training, the image with an output sentence that has a higher cosine similarity to the input caption than the other image will be chosen as the target. 

This module allows the agents to quickly and accurately make the decisions without further training. In theory, if the speaker and the listener were to have the exact same brain (same model and weights), the performance of this task should approach 100\%. The results of ``No Disparity'' speaker in Figure \ref{acc} confirmed the design choice.

\subsection{Rational Speaker ($S_1$)} \label{s1}
Without disparity concerns, the Rational Speaker ($S_1$) fulfills the \textbf{task goal} by simulating (Figure \ref{smodel}) the Rational Listener ($L_1$)'s behavior, and rank the candidate captions generated by the Literal Speaker ($S_0$) according to how well they can describe the target image apart from the distractors. This design is under the fair assumption that both speakers and listeners are aware of the collaborative game goal, but can be switched for other task purposes.

% \begin{equation}
% \begin{split}
%     S_1: P(c|m_0, m_1, t) := argmax_{c \in \mbox{Cand } C}\\ P_{L1}(t=t^*|m_0,m_1,c)
% \end{split}
% \end{equation}

\begin{equation}
\begin{split}
    &\mbox{For } i\in\{0, \cdots, n\}, \mbox{where } n = |C|:\\
    &t_i, p_i = \mathtt{Simulate\_L_1}(m_0, m_1, c_i)\\
    &c = c_{\mathrm{argmax}_{i}[[t_i == t^*]]\cdot p_i}
\end{split}
\end{equation}

Given an image pair ($m_0, m_1$), and a list of candidate captions $C = \{c_1, \cdots, c_n\}$ generated by $S_0$, the Rational Speaker goes through each caption $c_i$ and simulates how well the listener ($\mathtt{Simulate\_L_1}$) would pick out the correct target image. If a candidate caption $c_i$ helps the simulator pick out the correct target image (i.e. $t_i == t^*$) with high confidence ($p_i$), then it will be chosen as the final caption sent over to the actual listener. The simulated listener shares the same architecture as $L_1$ and initializes the weights pre-trained from $S_0$. By doing so, the Rational Speaker takes the listener's intention into account and achieves the task goal.

% The architecture of the simulated listener module is the same described in Section \ref{l1}. As the literal speaker generates a list of candidate captions, the working memory reranks each caption to decide 1) whether this caption would allow the listener to pick the correct target image, and if so 2) how confident, assuming no disparities. By doing so, the speaker can avoid sending over captions that can be used to describe both images, and help the listener to pick out the correct target image.

\subsection{Listener with Disparities ($L_1^d$)} \label{disp}
In the real world, however, it is hardly the case that different agents have the exact same knowledge background, experiences, physical capabilities, etc. The listener's decision making process is influenced by various kinds of disparities $d$.
% \begin{equation}
%     L_1^d: P(t|m_0, m_1, c, d)
% \end{equation}

% \begin{equation}
% \begin{split}
%  L_1^d: P(t|m_0, m_1, c, d) := argmin_{t\in\{0,1\}}\\ \mbox{CosSim}(\mbox{Transformer}^d(m_t, c), c)
% \end{split}
% \end{equation}

To study speaker's ability of situated language adjustment, we created two representative types of listeners with different knowledge background and visual capabilities by training different caption grounding modules ($\mathtt{FT\_Transformer}$) with the datasets assembled in Section \ref{dataset}. These disparities would challenge the speaker model to adjust the language at different levels.
% We chose these two types of disparities because they represent two distinctive types of perceptual adjustment challenges: lexical replacement, and semantic attention shift.

\begin{enumerate}
%\vspace{-3pt}
    \item $L_1^{d1}:$ Hypernym. With limited vocabulary and knowledge in a certain domain, people tend to refer to objects in their hypernym form (e.g. ``animal'' instead of ``cat''). In this experiment, we create listeners that would refer to all the detected objects by their hypernyms. This disparity would require the speaker to switch individual words that share similar meanings.
    \item $L_1^{d2}:$ Limited Visual. Due to the physical orientation or impaired vision capability, it is likely that some objects are blocked or hardly visible to one party but not the other. In this experiment, we remove all the animal objects from listener's visual detected object list ($O$), and replace the relevant descriptions with the special token `[UNK]'. This disparity would require the speaker to shift attention, and choose alternative objects to describe.
\end{enumerate}
%\vspace{-3pt}

We investigate in listeners with a subset of speaker's capabilities under the argument that in the opposite case, the listener could use only a subset of the knowledge to achieve best performance without having the speakers to adjust the speech. Other disparities can be inferred through transfer learning or are left for further investigation with broader information access and datasets.

\begin{figure*}[h]
     \centering
     \begin{subfigure}[b]{0.45\textwidth}
         \centering
         \includegraphics[width=0.98\textwidth]{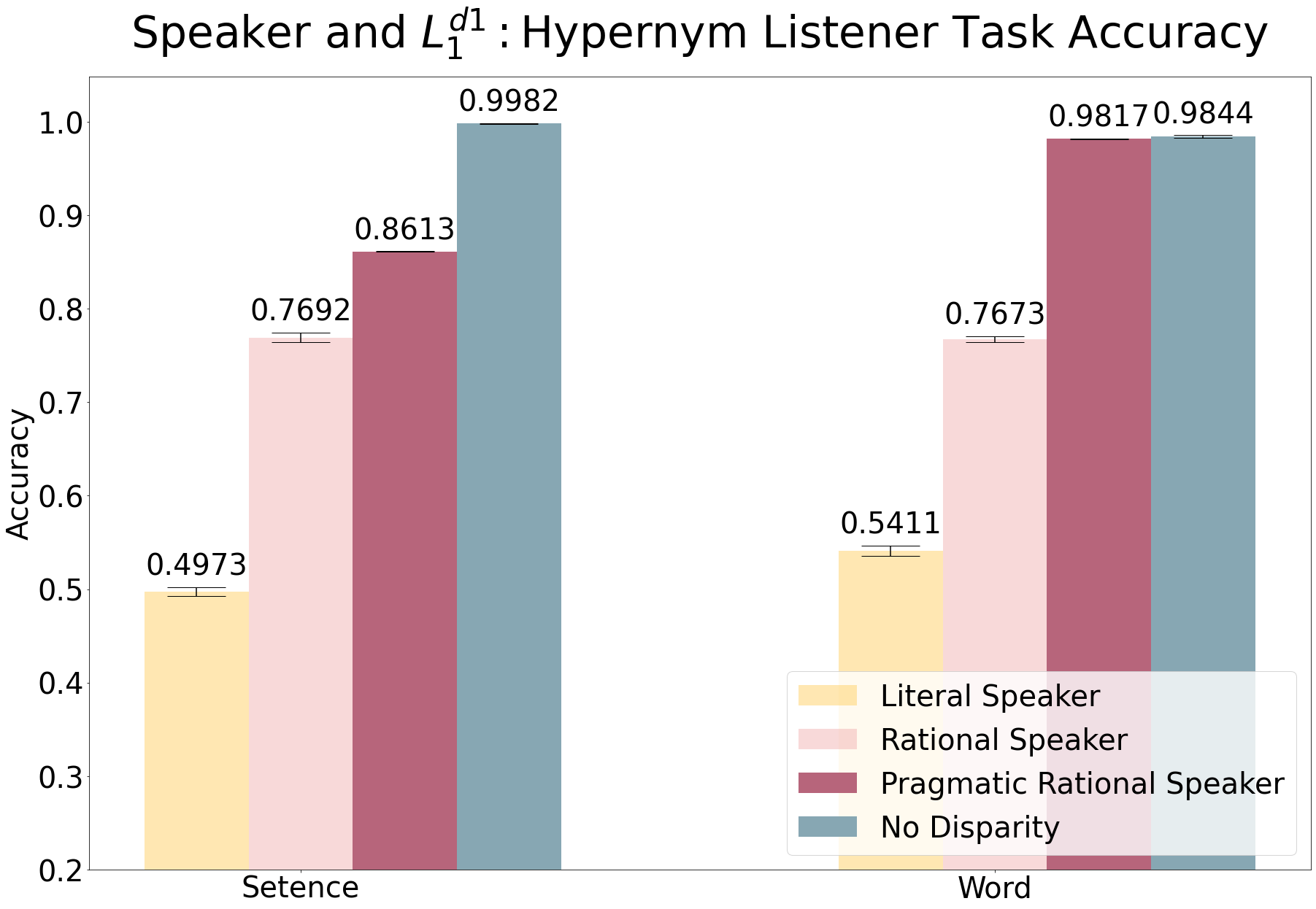}
         \caption{$L_1^{d1}$: Hypernym}
         \label{acc1}
     \end{subfigure}
     \hfill
     \begin{subfigure}[b]{0.45\textwidth}
         \centering
         \includegraphics[width=0.98\textwidth]{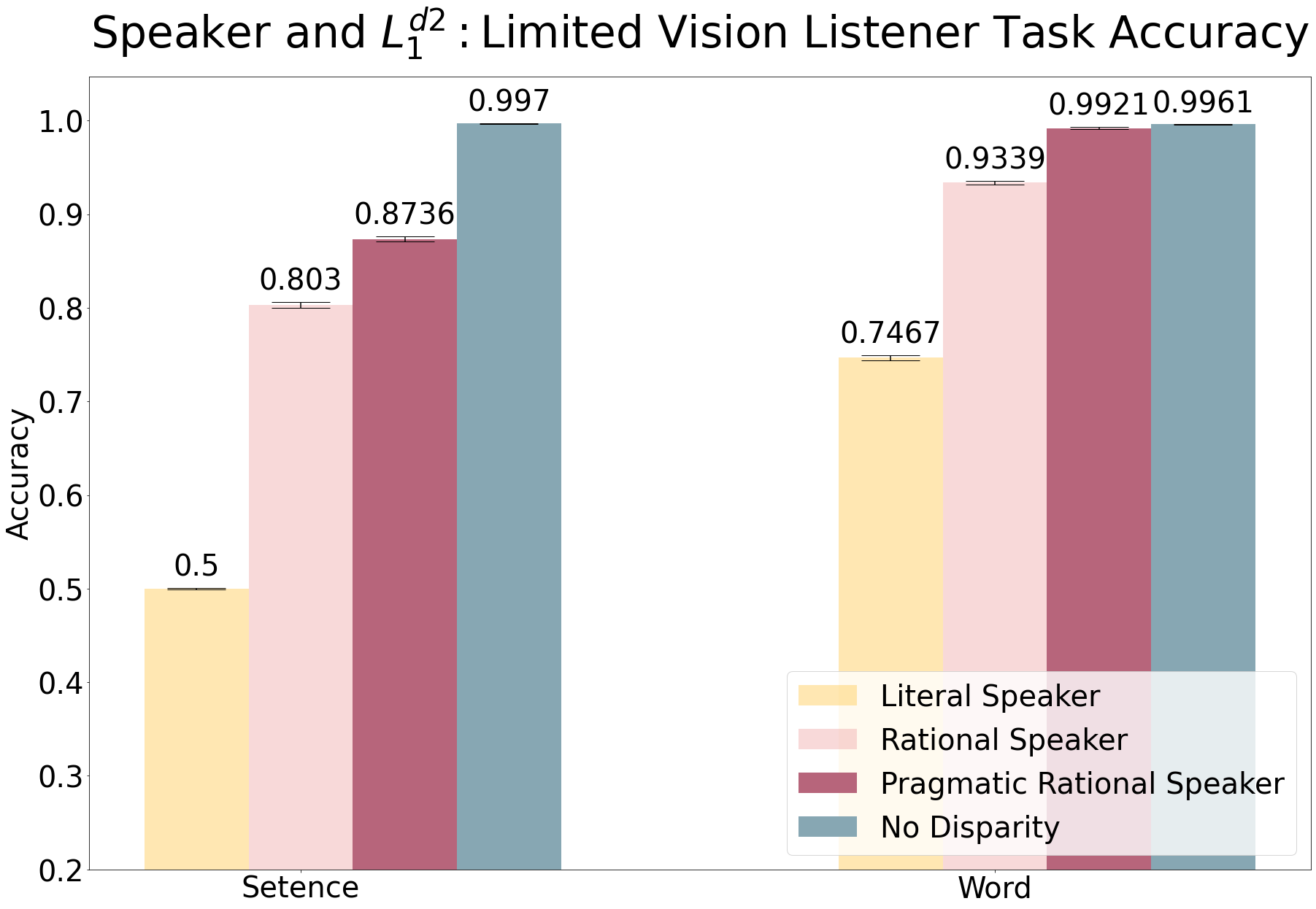}
         \caption{$L_1^{d2}$: Limited Vision}
         \label{acc2}
     \end{subfigure}
     %\vspace{-8pt}
     \caption{Referential game Accuracy: The Pragmatic Rational Speakers are able to significantly outperform Literal Speakers and vanilla Rational Speakers across different types of disparities. Word level models achieve higher performance and is much closer to the No Disparity upper bound than the sentence level communication.}
     \vspace{-8pt}
     \label{acc}
\end{figure*}

% \begin{table*}[]
%     \centering
%     \begin{tabular}{|c|l|c|c||c|c|}
%     \hline
%     %  & $L_1^{d1}$: Hypernym &&& 
%     %     & $L_1^{d2}$: Limited Vision &&& \\
%     %  {\multirow{2}{*}{}}&{\multirow{2}{*}{Pragmatic Rational Speaker}}& \multicolumn{2}{|c||}{$L_1^{d1}$: Hypernym} & \multicolumn{2}{|c|}{$L_1^{d2}$: Limited Vision} \\\cline{3-6} 
%     \multicolumn{2}{|c|}{\multirow{2}{*}{Pragmatic Rational Speaker}} & 
%     \multicolumn{2}{|c||}{$L_1^{d1}$: Hypernym} & \multicolumn{2}{|c|}{$L_1^{d2}$: Limited Vision} \\\cline{3-6} 
%      \multicolumn{2}{|c|}{} & Hard & Easy 
%         & Hard & Easy\\ \hline
%      {\multirow{2}{*}{Sentence}} & Compared to Rational Speaker & +11.35 &  +13.64 
%         & +7.95	& +9.60\\ 
%      & Compared to No Disparity & -20.26 & -9.64 
%         & -19.29 & -9.09\\ \hline
%      {\multirow{2}{*}{Word}} & Compared to Rational Speaker & +19.27	&	+27.37
%         & +2.00	& +6.28\\
%      &  Compared to No Disparity & -5.90	& -0.85
%         & -5.15	&-0.00\\ \hline
     
%     \end{tabular}
%     \caption{Pragmatic Rational Speaker's Performance (Absolute Percentage Change) Improvement Comparison}
%     \vspace{-10pt}
%     \label{diff}
% \end{table*}

\subsection{Pragmatic Rational Speaker ($S_1^d$)} \label{s1d}
On top of the Rational Speaker ($S_1$), the Pragmatic Rational Speaker incorporates a disparity adjustment layer to learn and accommodate the listener's disparity through emergent communication. 

% \begin{equation}
% \begin{split}
%     S_1^d: P(c|m_0, m_1, t, d) \\
% := argmax_{c\in \mbox{Cand }C} \mbox{Reward}(\mbox{MLP}(\mbox{Emb}(c)))
% \end{split}
% \end{equation}

\begin{equation}\label{eq_s1d_2}
\begin{split}
&\mbox{For } i\in\{0, \cdots, n\}, \mbox{where } n = |C|:\\
&q_i = \mathtt{MLP}(\mathtt{SentenceEmb(c_i)})\\
&a_i = [[t_i == t^*]]\cdot p_i\cdot q_i\\
&c = c_{\mathrm{argmax}_{i} a_i}\\
%&c = \mathrm{argmax}_{c_i\in C}[[t_i == t^*]]\cdot p_i\cdot q_i
\end{split}
\end{equation}

We use a pretrained BERT model to embed each candidate caption $c_i$, add a single MLP layer, and approximate the REINFORCE policy through Equation \ref{eq_s1d_2}. The reward ($r_{c^*}$) for each chosen caption $c^*$ is $+1$ or $-1$. The loss is calculated for all the chosen captions across each batch (Eq \ref{eq_loss}).
% We model this by embedding each candidate captions with pretrained BERT embeddings, adding a single Linear layer, and approximating the REINFORCE policy according to the equation below. The reward is $+1$ when the listener picks out the correct target image, and $-1$ otherwise. 

%\vspace{-5pt}
\begin{equation}\label{eq_loss}
    L = -\sum_{c^*} \mbox{log}(a_{c^*})\cdot r_{c^*}
\end{equation}
%\vspace{-20pt}

% \begin{equation}
% \begin{split}
%     S_1^d: P(c|m_0, m_1, t, d)\propto \\
% P(c|m_0, m_1, t)^{\lambda_l} \cdot P(t|m_0, m_1, c, d)^{\lambda_d} \\
% := argmax_{c\in \mbox{Cand }C} \mbox{Reward}(\mbox{MLP}(\mbox{Emb}(c)))
% \end{split}
% \end{equation}

% The working memory of the Pragmatic Rational Speaker ($S_1^d$) has two two goals: 1) \textbf{Task Goal}: an internal simulation of a listener to rank the candidate captions by their uniqueness in describing only the target image, and 2) \textbf{Disparity Goal}: a disparity adjustment layer to learn and accommodate the listener's disparity through reinforcement interactions. Each goal component can be formalized in the above two terms. We parameterized each term with $\lambda_l$ and $\lambda_d$ to study how different $\lambda_l : \lambda_d$ weight ration could affect rational speaker's ability to achieve both goals.

\subsection{Communication with Words} \label{wd}
We conducted the same sets of experiments using individual words (object names) instead of sentences to demonstrate the effects of working memory on disparity accommodation and internal task simulation, reducing the noise that came from the imperfection of the image description generator. The simplified pipeline uses the detected object name embedding for disparity adjustment, and the listener picks the target images by conducting simple word matching.

\section{Results and Analysis}

We evaluate our models ($S_0, S_1, S_1^d$) on the referential game (Figure \ref{imgexp}) along four dimensions: {\bf{End-task Performance}, \bf{Efficiency}, \bf{Transparency}, and \bf{Balance of Goals}.} Recall that each speaker model has different capabilities (Table \ref{sl}) and only $S_1^d$ is able to fulfill both \textit{task} and \textit{disparity} goals. Implementation details and more experiment results can be found in Appendix.

\begin{enumerate}
    %\vspace{-5pt}
    \item {\bf [Task Performance]} that measures overall accuracy of the collaborative game. Task performance is often the sole evaluation metrics in previous work.
    %\vspace{-5pt}
    \item {\bf [Efficiency]} that measures time used for model training across tasks.
    %\vspace{-5pt}
    \item {\bf [Transparency]} that uncovers the underlying distribution shift of vocabulary use learned to accommodate different types of disparities. 
    \item {\bf [Balance of Goals]} that the working memory needs to consider between the task and disparity goals to achieve maximum performance 
    %\vspace{-5pt}
\end{enumerate}

% We introduce two additional baselines for comparison: Joint Trained Speaker (Section \ref{sec_eff}) to demonstrate the benefits of working memory and separate training, and No Disparity (Section \ref{perfm}) to showcase the limitations of our models. More qualitative examples and additional experiment results can be found in the Appendix.

\subsection{Task Performance Comparison}\label{perfm}

% \begin{figure*}[t]
%      \centering
%      %\begin{subfigure}[b]{0.45\textwidth}
%          \centering
%          \includegraphics[width=0.98\textwidth]{figs/shift_l1.png}
%          \caption{$L_1^{d1}$: Hypernym}
%   \label{acc}
% \end{figure*}

To assess the performance of the speakers in the collaborative game, Figure \ref{acc} presents the task accuracies with Literal Speaker ($S_0$), Rational Speaker ($S_1$), Pragmatic Rational Speaker ($S_1^d$), and No Disparity ($S_1^{nd}$). $S_1^{nd}$ has the same structure as $S_1$ and was trained on the same disparity dataset as the corresponding listener. It serves as the upper bound of performance. The same experiments also were conducted at the word level.

For each type of listener disparity, the performance is $S_0 << S_1 < S_1^d < S_1^{nd}$. The vanilla Rational Speaker ($S_1$) improved the overall performance from Literal Speaker by over 25\% because it is achieving the task goal to describe the target image apart from the distractor. The Pragmatic Rational Speaker ($S_1^d$) is able to learn and adjust for the listener's disparity, and further improve the game performance by $\sim$10\%. There is still, however a gap between $S_1^d$ and the upper bound $S_1^{nd}$, where the speaker and the listener have the exact knowledge and capability limitation, potentially due to the imperfection in caption generations.

\begin{figure}[h]
     \centering
     \includegraphics[width=0.48\textwidth]{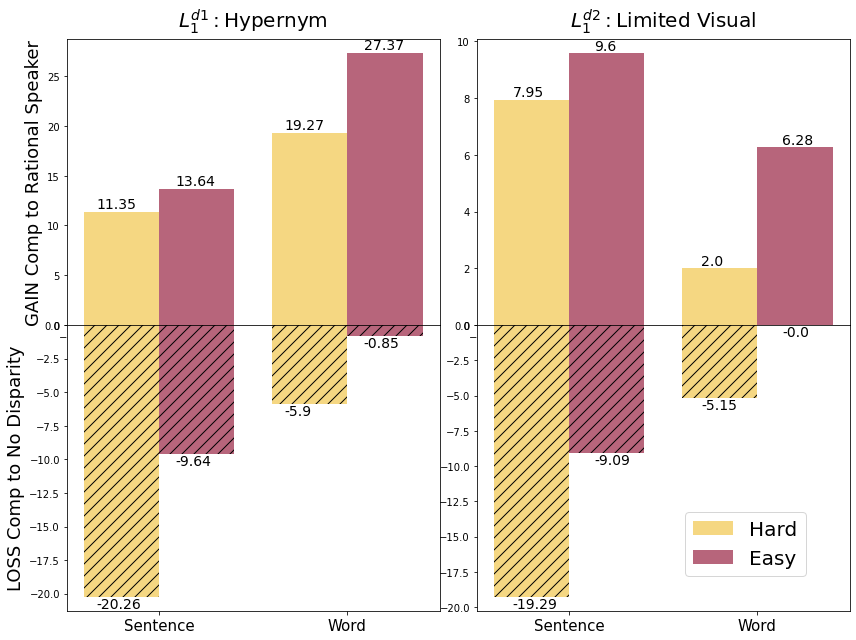}
     %\vspace{-15pt}
    \caption{Pragmatic Rational Speaker Performance Gain/Loss compared to Rational Speaker and No Disparity (upper bound).}
     %\vspace{-15pt}
     \label{gain}
\end{figure}

Breaking down between the \textit{hard}, \textit{easy} datasets in Figure \ref{gain} (recall that image pairs that differ by $\leq4$ objects are in the \textit{Hard} set, otherwise the \textit{Easy} set), $S_1^d$ on the \textit{easy} dataset is able to gain a lot more improvement upon its Rational Speaker compared to the pair trained on the \textit{hard} dataset. The gap between $S_1^d$ and No Disparity is also a lot smaller for the model trained on the \textit{easy} dataset. This is likely because when a pair of images differ more objects (easier), the model has more options to adjust upon, hence the larger improvement. 

Compared to the sentence level model, the word level pragmatic speaker for $L_1^{d1}$ achieves even higher improvement against the corresponding Rational Speaker. They both achieve almost perfect accuracy with close to zero gap to the upper bound. This suggests the high potential of the disparity adjustment design, especially after reducing the caption generation and interpretation noise.

\begin{table}
    \begin{subtable}[h]{0.45\textwidth}
        \centering 
        \begin{tabular}{ |l  l | c | c | c|} \hline
        &&Train(min) &  Accuracy\% & BLEU4\\ \hline
        \multicolumn{2}{|l|}{Joint} & 19.04 & 60.14 & 27.79 \\ \hline
        \multicolumn{2}{|l|}{Separate} & 21.02 & \textbf{77.34} &	\textbf{29.3} \\ \hdashline
        &$\bullet$ LM &	11.59	&&	29.3 \\
        &$\bullet$ WM &	\textbf{9.43}&	77.34& \\ \hline
      \end{tabular}
      \caption{$L_1^{d1}$: Hypernym}
      \label{eff_l1}
    \end{subtable}
    \hfill
    \begin{subtable}[h]{0.45\textwidth}
        \centering 
        \begin{tabular}{ |l  l | c | c | c|} \hline
        &&Train(min) &  Accuracy\% & BLEU4\\ \hline
        \multicolumn{2}{|l|}{Joint} & 29.52&	63.69&	27.29 \\ \hline
        \multicolumn{2}{|l|}{Separate} & 29.95	&\textbf{81.09} &	\textbf{29.3} \\  \hdashline
        &$\bullet$ LM &	11.59	&&	29.3 \\
        &$\bullet$ WM &	\textbf{18.36} &	81.09& \\ \hline
      \end{tabular}
      \caption{$L_1^{d2}$: Limited Vision}
      \label{eff_l2}
    \end{subtable}
    %\vspace{-5pt}
     \caption{Compared to joint training, separate training only needs to train the long-term memory once, and can achieve higher performance. \textbf{LM}: Long-term Memory, \textbf{WM}: Working Memory.}
     %\vspace{-15pt}
     \label{eff}
\end{table}

\begin{figure*}[t]
     \centering
     \hfill 
     \begin{subfigure}[b]{\textwidth}
        \centering
         \includegraphics[width=0.98\textwidth]{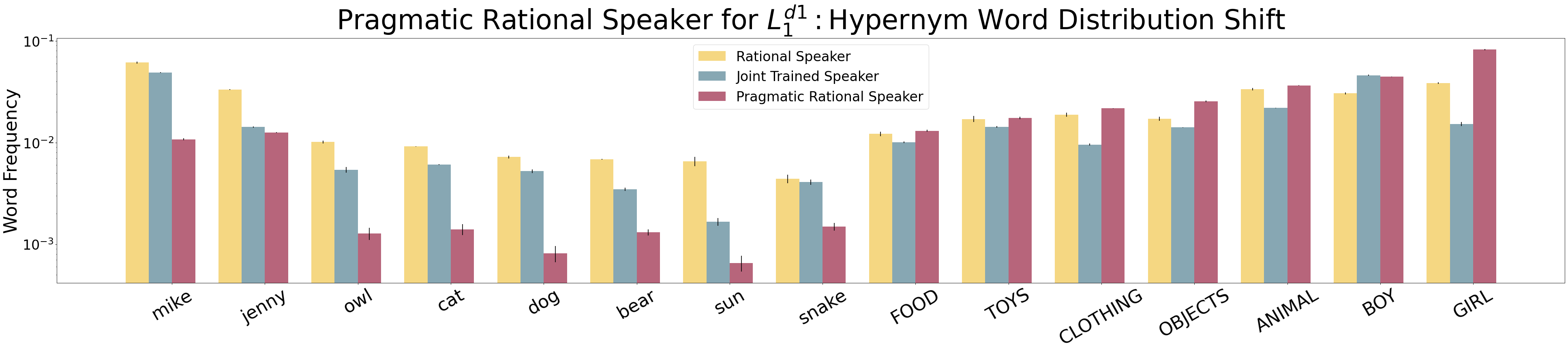}
         %\vspace{-10pt}
         \caption{$L_1^{d1}$: Hypernym}
         \label{shift_h}
     \end{subfigure}
     \hfill
     \begin{subfigure}[b]{\textwidth}
         \centering
         \includegraphics[width=0.98\textwidth]{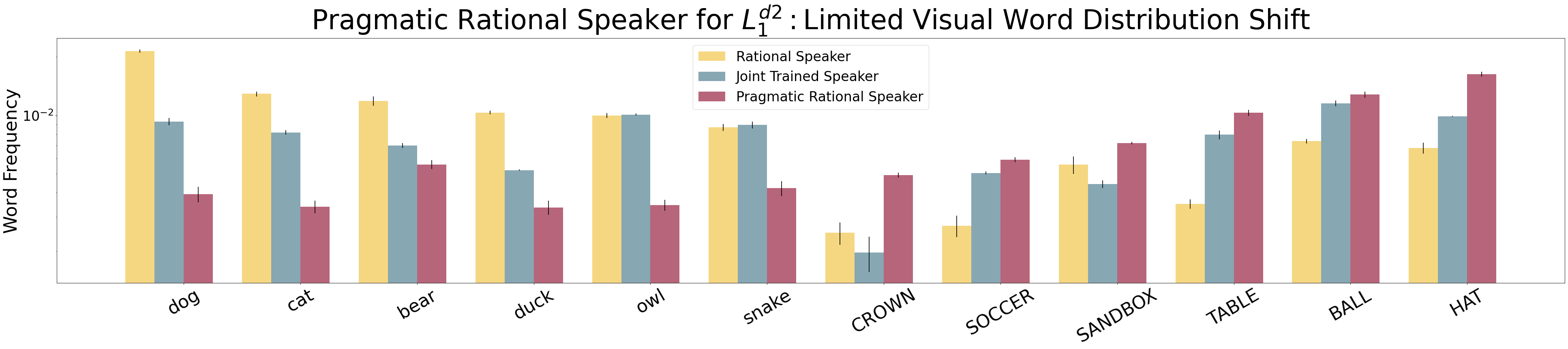}
         %\vspace{-10pt}
         \caption{$L_1^{d2}$: Limited Vision}
         \label{shift_v}
     \end{subfigure}
     %\vspace{-20pt}
          \caption{Word Distribution Shift: The Pragmatic Rational Speaker for $L_1^{d1}$ avoids specific object names and prioritizes some hypernyms. The Pragmatic Rational Speaker for $L_1^{d2}$ avoids animal related words in communication.}
          \vspace{-10pt}
     \label{shift}
\end{figure*}

\begin{figure}
    %\vspace{-20pt}
     \centering
     \begin{subfigure}[b]{0.45\textwidth}
         \centering
         \includegraphics[width=0.98\textwidth]{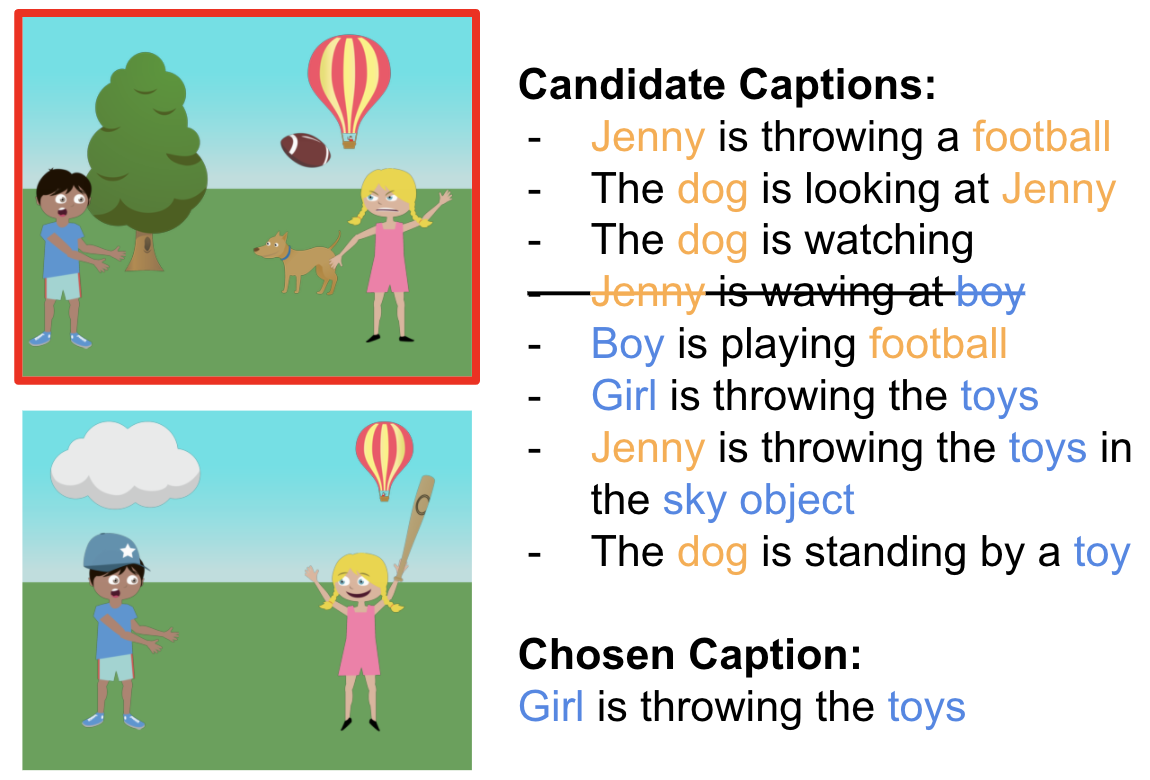}
         \caption{$L_1^{d1}$: Hypernym}
     \end{subfigure}
     \hfill
     \begin{subfigure}[b]{0.45\textwidth}
         \centering
         \includegraphics[width=0.98\textwidth]{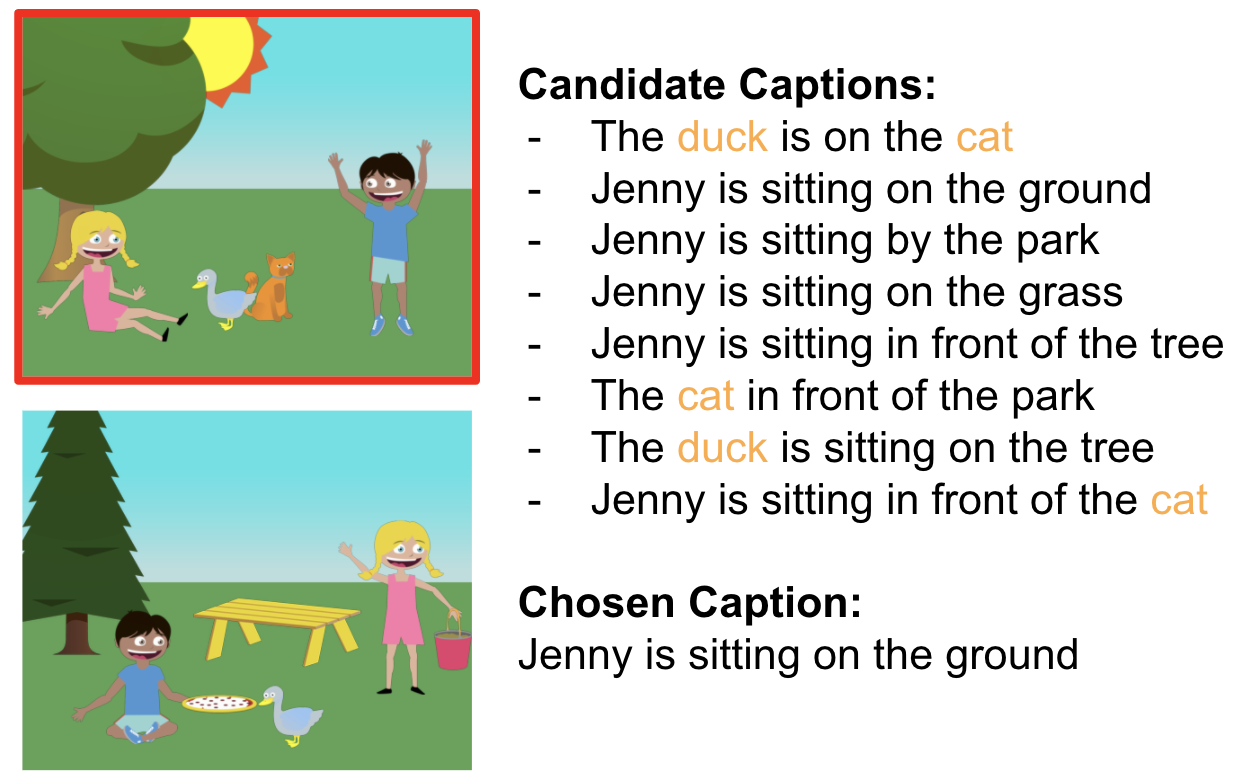}
         \caption{$L_1^{d2}$: Limited Vision}
     \end{subfigure}
   % \vspace{-8pt}
    \caption{ Qualitative examples for each disparity adjustment. The \textcolor{orange}{orange} words are the vocabulary that the Pragmatic Rational Speaker is avoiding, and the \textcolor{blue}{blue} words are the preferred alternative for the listeners. The \sout{strikethrough} sentences are discarded because they can be used to describe both images. }
    %\vspace{-10pt}
     \label{qualit}
\end{figure}

\vspace{-10pt}
\subsection{Learning Efficiency} \label{sec_eff}

% \jycc{It would be more informative if we have a better way to demonstrate efficiency. Here you want to say that joint training will need to train again and again for every task, but our approach only needs to train working memory. So which is a lot more efficient. I'm wondering if we can test on a combination of tasks, and show the aggregated training time (total) across all tasks. Currently, the table 4 does not demonstrate efficiency}

To study the training efficiency of the working memory, we compared our model to the joint training ``Multi-Task leaning'' model in \cite{lazaridou}'s work, retrained and evaluated in our dataset. The image captioning model and the REINFORCE learning are joint trained through a combined loss:
$$L = \lambda_f L^{functional} + \lambda_s L^{structural}$$
\textit{Functional} in our task refers to the REINFORCE learning to achieve both task and disparity goals (evaluated by Accuracy), and  \textit{structural} refers to the caption generation loss for natural-sounding language (evaluated by BLEU4). We used $\lambda_f = \lambda_s = 1$ as in previous work for our experiments.

% \begin{table*}[]
%     \centering
%     \begin{tabular}{|l|l|c|c|c|} \hline
%          & Disparity & Train(min) & Accuracy & BLEU4\\ \hline
%          \textbf{Long-term Mem} & & 11.59 & & \textbf{29.30}\\ \hline \hline
%          Joint Train & $L_1^{d1}$: Hypernym & 19.04 & 60.14 & 27.79\\ 
%          \textbf{Working Mem} & $L_1^{d1}$: Hypernym & 9.43 & \textbf{77.34} & \\ \hline \hline
%          Joint Train & $L_1^{d2}$: Limited Vision & 29.52 & 63.69 & 27.29 \\ 
%          \textbf{Working Mem} & $L_1^{d2}$: Limited Vision & 18.36 & \textbf{81.09} & \\ \hline \hline
%          Joint Train & Both & 48.56 & &\\ 
%          \textbf{Sep Train} & Both & \textbf{39.38} & &\\ \hline
%     \end{tabular}
%     \caption{Compared to joint training, our method is able to train much faster for multiple types of listener disparity learning by fixing the long-term memory and only train the working memory flexibly for each type. It also achieves higher performance in both accuracy and BLEU4 score per disparity.}
%     \label{eff}
% \end{table*}

Detailed training and comparison strategies can be found in the Appendix. Table \ref{eff} shows that for each type of disparity, our model separating working memory from long-term memory is able to achieve higher accuracy and higher BLEU4 score than the joint training. Moreover, the Joint Trained model needs to retrain all the weights for each type of disparity from scratch, whereas our model only needs to train the long-term memory once, and retrain the light weighted working memory for each type of disparity, which is much more efficient.

\subsection{Transparency: Vocabulary Adjustment}

To gain insights in whether the Pragmatic Rational Speaker (PRS) is actually adjusting the descriptions for listeners' disparities or taking the advantage of statistical bias to achieve higher task performance, we plotted the word distribution shift across different types of disparities. Qualitative examples can be found in Figure \ref{qualit}. For each experiment, the word frequencies of all the chosen captions were calculated for the Rational Speakers, the Pragmatic Rational Speakers, and Joint Training. We collected the top choice of each speaker per image pair, repeated the experiments 3 times, and reported the mean and standard deviation in Figure \ref{shift}.

In the Hypernym disparity (Figure \ref{shift_h}) experiment, where the listener only understands the hypernym of detected objects, the lower-case words on the left are the top detected object names, and the upper-case words on the right are hypernyms. On the left side, the word frequencies of PRS significantly dropped from the Rational Speaker. On the right side, the model is maintaining similar level, or using some of the hypernyms more frequently (y-axis in log scale). Note that the Rational Speaker can generate both hypernym and hyponym regardless of disparities, and multiple valid captions available for all speakers to choose from. For the Joint Trained Speaker, we also observed a hyponym usage drop (left), but it's unclear how it accommodates the disparity without using hypernyms. This result shows that PRS learned to avoid using hyponyms, and replaced them with their hypernym to accommodate the disparity.

% the overall frequency mass of the Pragmatic Rational Speaker is higher than than of the Rational Speaker, with most of sentences switched to only talking about ``Boy'' and ``Girl'' (Note: y-axis in log scale). This shows that the model learned to avoid the specific object names, and prioritize some hypernyms to accommodate the $L_1^{d1}$ disparity. 

For the Limited Visual disparity (Figure \ref{shift_v}), since all the animal objects are missing for the listener, there is a sharp decline in $S_1^{d2}$'s use of animal related words during the communication. Instead, it is choosing other objects such as ``hat'', and ``ball'' to describe the target image. The PRS is accommodating listener $L_1^{d2}$'s disparity by shifting the attention and choosing alternative objects other than animals to communicate. The behavior of the Joint Trained Speaker is harder to interpret.

\subsection{Balancing Between Goals}

Recall that the working memory of the Pragmatic Rational Speaker ($S_1^d$) has two two goals: 1) \textbf{Task Goal}: an internal simulation of a listener to rank the candidate captions by their uniqueness in describing the target image, and 2) \textbf{Disparity Goal}: a disparity adjustment layer to learn and accommodate the listener's disparity through interactions. Each goal component can be formalized in the above two terms (Equation \ref{twog}). We parameterized each term with $\lambda_l$ and $\lambda_d$ to study how different $\lambda_l : \lambda_d$ weight ratio could affect rational speaker's ability to achieve both goals.

\begin{equation}
\begin{split}
&a_i = ([[t_i == t^*]]\cdot p_i)^{\lambda_l} \cdot (q_i)^{\lambda_d}\\
\end{split}\label{twog}
\end{equation}

\begin{figure}[h]
     \centering
     \includegraphics[width=0.45\textwidth]{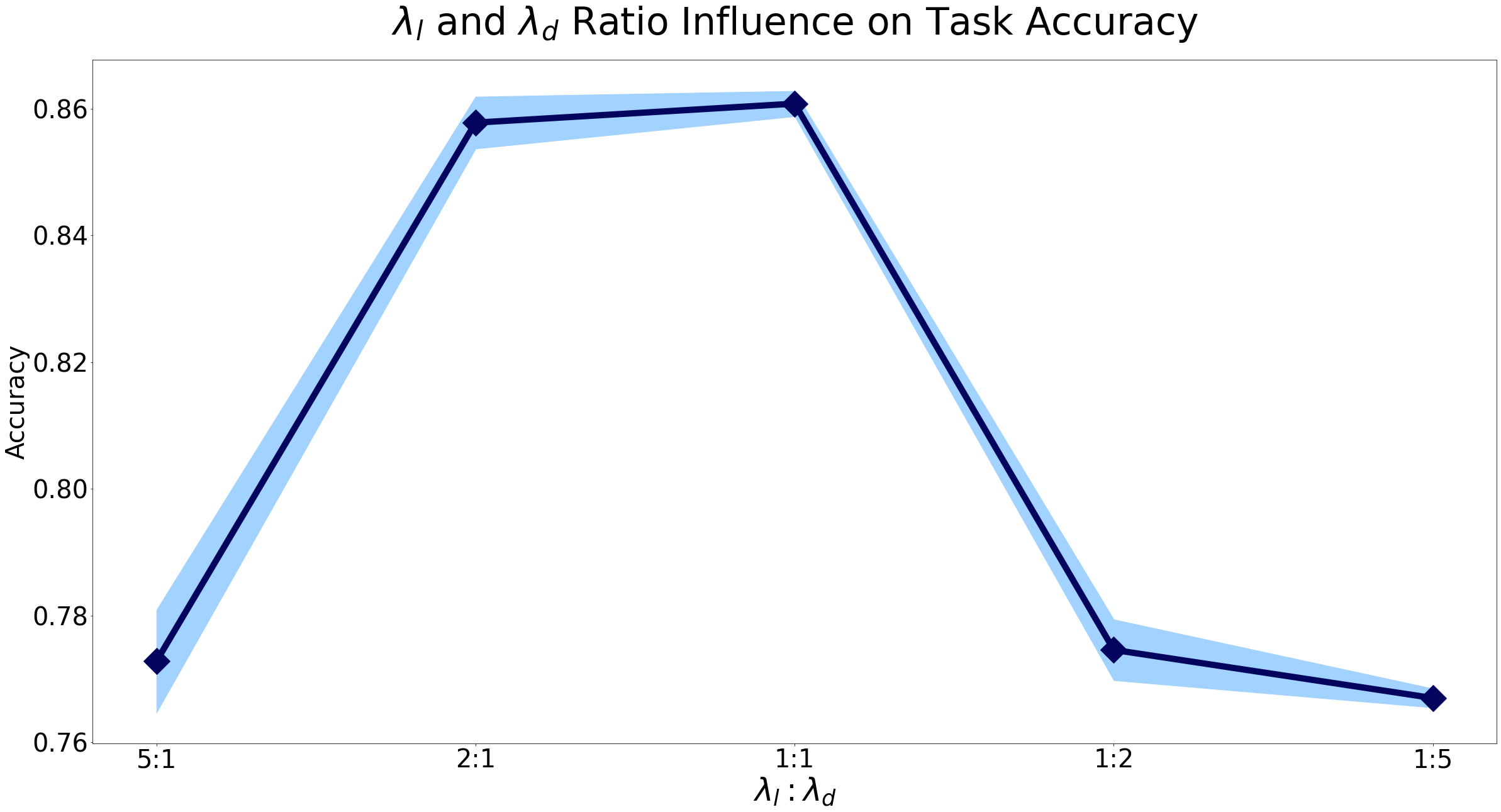}
     \caption{Balancing between the task goal and the disparity adjustment goal: the Pragmatic Rational Speaker needs a balanced emphasise on both $\lambda_l$ and $\lambda_d$ in order to achieve both goals simultaneously. }
     \label{lambda}
\end{figure}

Figure \ref{lambda} shows that when the Pragmatic Rational Speaker puts a high emphasis on adjusting the listener's disparity $\lambda_d$, it would ``forget'' to describe the unique characters of the target image and lower the overall performance. On the other hand when the PRS emphasize too much on the task goal, it would ``forget'' to accommodate listener's disparities, and lower the overall performance as well. In the end, we chose $\lambda_l : \lambda_d = 1:1$ for all experiments demonstrated above.

\section{Conclusion and Future Work}
% In this work, we extended the Rational Speech Act framework in the multi-agent communication task to accommodate the agent knowledge background disparities. The novel application of the light-weighted working memory can efficiently adjust the long-term memory for the corresponding listener. The introduction of an internal simulation module also demonstrated a strong performance in the task-specific decision making process. 
In this work, we present a novel framework based on the Rational Speech Act framework for pragmatic communication that can adjust the conversation content according to listener's disparities by adding a light-weighted working memory on top of speaker's long-term memory. The Pragmatic Rational Speaker significantly improved the collaborative game performance by shifting the conversation towards the language that the listeners are able to understand. The flexibility and training efficiency also makes it easy to be applied broadly.

There are, however, several limitations that requires further investigation. 
%One of the limitations is the image captioning module, and therefore, the literal speaker and listener models. More specifically, caption generation doesn't 100\% resemble the image due to complicated object interactions, and the top candidates may not cover all the word switches. Pragmatic inference models heavily relies on great functioning base models. 
First of all, despite recent progress,  algorithms that connect language and the visual world are still limited. For example, caption generation, even in this simple setup, often does not faithfully capture what's been conveyed in the images. As our framework heavily relies on the quality of various models that bridge language and vision, e.g., as part of our long term memory, it's important to improve functionality and performance of these base models. 
%, whether it's language generation from visual information or . 

We conducted our experiments in a relative simple and artificial environment with the purpose of easy control and demonstration. We emphasize on evaluating model's actual language ability of adjusting for the disparities on top of task performance. The next step would be to apply the framework to more realistic images and interactive environment. 

Other than listener's knowledge background and perceptual capabilities, there are a lot of other reasons for language communication to be adjusted, such as the physical environment, relative positions, speaker's personalities, etc. Studying how a rational agent can accommodate these disparities would require additional multimodal datasets and information processing methods. 

% Last but not least, it would be great for pragmatic reasoning to meta-learn and zero-shot transfer existing knowledge to different disparities and situations. Future work could situate the model in more complex and realistic settings with different disparity types, and improve for more general and faster learning.

% One of the limitations is the image captioning module, which doesn't 100\% resemble the image due to complicated object interactions, and the top choices may not cover all the word switches. 
% % Moreover, a simple concatenation of the label embedding and bounding box might not be sufficient to specify the relative locations of all the objects, and hence their interactions. 
% Pragmatic inference models relies great functioning base models, and the performance of a base model could affect the outcome of the pragmatic model.

% We conducted our experiments in a relative simple and artificial environment with the purpose of easy control and demonstration. We emphasize on evaluating model's actual language ability of adjusting for the disparities on top of task performance. The next step would be to apply the framework to more realistic images and interactive environment.

% Other than listener's capabilities, there are a lot of reasons for a conversation to be adjusted, such as the physical environment, relative positions, speaker's personalities, etc. Studying how a rational agent can accommodate these disparities would require additional multimodal datasets, and information processing methods.

At the moment, the Pragmatic Rational Speaker trains a new layer in working memory from scratch for each type of disparity. This could have backward influence on the long-term memory. In life-long learning \cite{lifelong} like humans, the working memory can shape their long-term memory. At the very least, the model could store each learned disparity adjustments for future encounter. This modification is left for future work.

Last but not least, instead of training for every single type of disparity to name, human learners have the ability of meta-learning and zero-shot transferring existing knowledge to a new category. Future work on pragmatic reasoning should be easily adaptable to different disparities and situations.

% \section{Conclusion}
% We present a novel framework based on the Rational Speech Act framework for multi-agent communication that can adjust conversation content according to listener's disparities by adding a light-weighted working memory on top of speaker's long term memory system. The pragmatic rational speaker significantly improved the collaborative game performance by shifting the conversation towards the language that the listeners are able to understand. The flexibility and training efficiency also makes it easy to be applied broadly. We believe this work can be greatly helpful for future situated language generation and adjustment.

%%%%%%%%%%%%%%%%%%%%%%%%%%%%%%%%%%%%%%%%%%%%%%%%%%%%%%%%%%%%%%%%
%%%%%%%%%%%%%%%%%%%%%%%%%%%%%%%%%%%%%%%%%%%%%%%%%%%%%%%%%%%%%%%%
%%%%%%%%%%%%%%%%%%%%%%%%%%%%%%%%%%%%%%%%%%%%%%%%%%%%%%%%%%%%%%%%
%%%%%%%%%%%%%%%%%%%%%%%%%%%%%%%%%%%%%%%%%%%%%%%%%%%%%%%%%%%%%%%%

\section*{Acknowledgements}
This work was supported in part by the National Science
Foundation under grant IIS-1949634. The authors
would like to thank the anonymous reviewers for
their valuable comments and suggestions.

\clearpage

\bibliography{anthology} %custom
\bibliographystyle{acl_natbib}

\clearpage
\appendix

\section{Speaker and Listener Model Architecture Breakdown}
\begin{figure}
        \centering
        \begin{subfigure}[b]{0.475\textwidth}
            \centering
            \includegraphics[width=\textwidth]{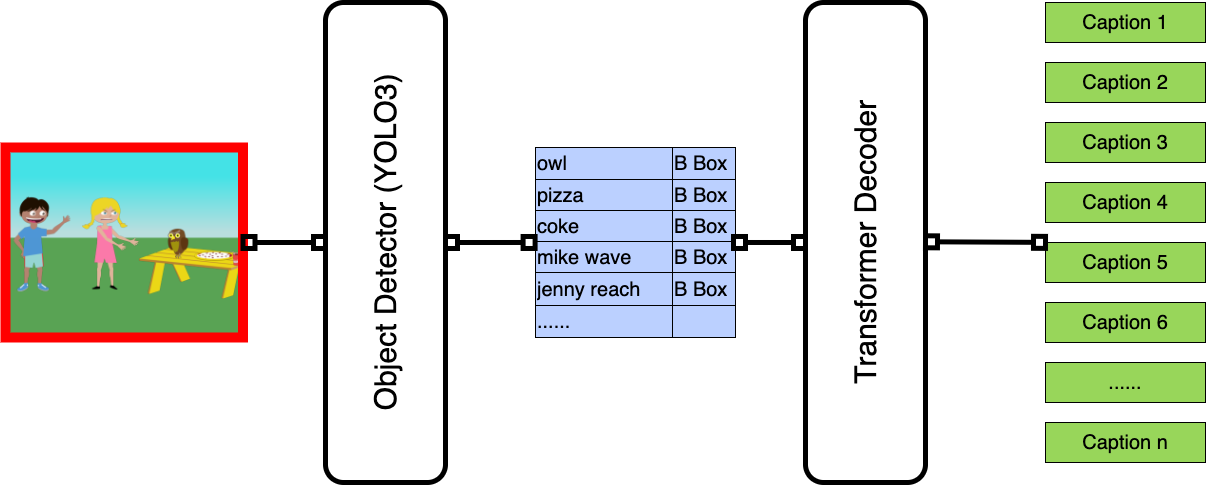}
            \caption[]%
            {{\small Literal Speaker $S_0$}}    
            \label{m_s0}
        \end{subfigure}
        \hfill
        \begin{subfigure}[b]{0.475\textwidth}   
            \centering 
            \includegraphics[width=\textwidth]{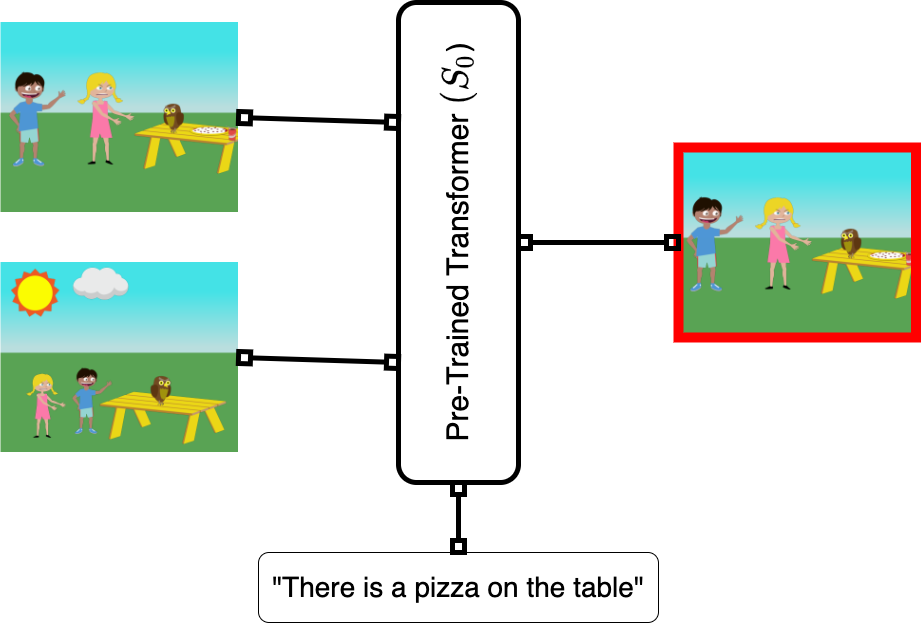}
            \caption[]%
            {{\small Rational Listener $L_1$/$L_1^d$}}    
            \label{m_l1}
        \end{subfigure}
        \hfill
        \vskip\baselineskip
        \begin{subfigure}[b]{0.475\textwidth}  
            \centering 
            \includegraphics[width=0.98\textwidth]{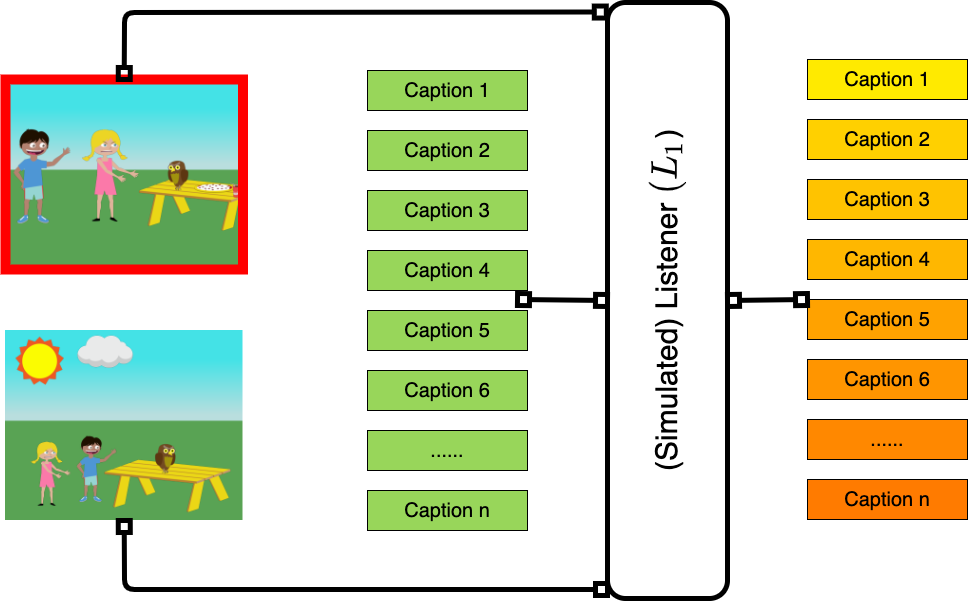}
            \caption[]%
            {{\small Rational Speaker $S_1$}}    
            \label{m_s1}
        \end{subfigure}
        \hfill
        \begin{subfigure}[b]{0.475\textwidth}   
            \centering 
            \includegraphics[width=0.98\textwidth]{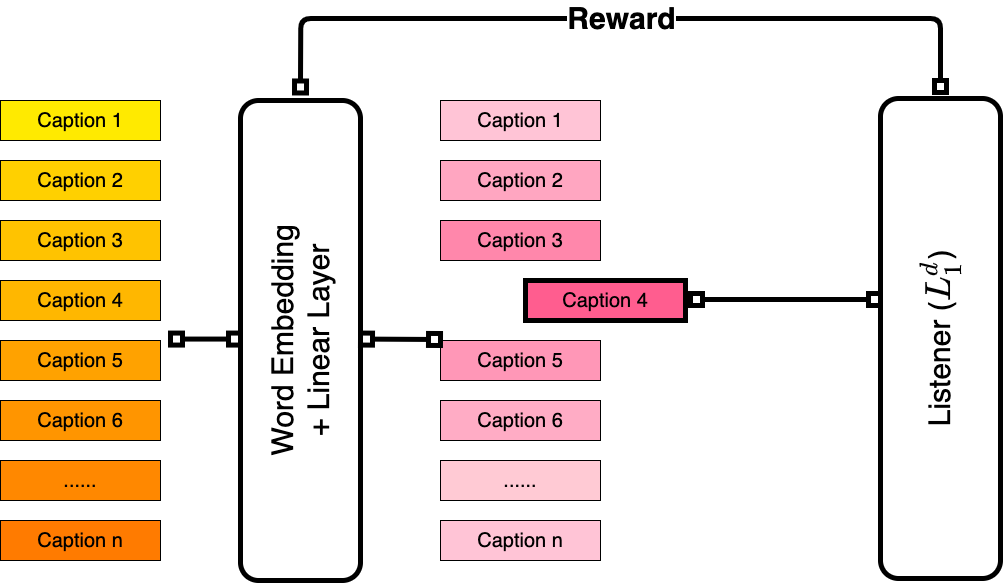}
            \caption[]%
            {{\small Pragmatic Rational Speaker $S_1^d$}}    
            \label{m_s1d}
        \end{subfigure}
        \hfill
        \caption[]
        {Speaker and Listener model breakdown} 
        \label{model_breakdown}
    \end{figure}

\begin{enumerate}[label=(\alph*)]
    \item \textbf{Literal Speaker $S_0$}: for each input image, we run YOLO3 object detector to get a list of detected object names. Each name is embedded with pre-trained BERT embedding, and concatenated with their bounding box location. The embedded images goes through a Transformer Decoder to generate a list of candidate captions.
    
    \item \textbf{Rational Listener $L_1$}: for each pair of images and an input caption, the Rational Listener reuses a pre-trained Transformer Decoder as in $S_0$ to figure out which image does the caption ground better in. Inspired the teacher-forcing caption training procedure, given an image and the input caption, if it generates a sentence that's closer to the input caption than the other image, then this image is chosen as the target.
    
    \item \textbf{Rational Speaker $S_1$}: for a pair of images, and a list of candidate captions generated by $S_0$, the Rational Speaker goes through each candidate caption via the internal simulated listener (same model as $L_1$ with no disparity), to figure out whether the caption can help the listener pick the correct target image, and if so, how confident. It ranks all the captions by the correctness and confidence score.
    
    \item \textbf{Pragmatic Rational Speaker $S_1^d$}: given a list of ranked (by $S_1$) candidate captions, the Pragmatic Rational Speaker picks the most confident one and send it to the actual listener with disparities ($L_1^d$), and receives a reward feedback. This feedback helps $S_1^d$ to learn the disparity, and rerank all the captions to accommodate the difference and optimize for the task goal.
    
\end{enumerate}

\clearpage

\begin{table*}

\begin{center}
\begin{tabular}{ |c|c||c|c| } 
\hline
Hypernym & Object & Hypernym & Object \\
\hline \hline
\multirow{7}{4em}{boy} & mike\_reach &\multirow{7}{4em}{girl} & jenny\_reach\\ 
& mike\_kick & & jenny\_kick\\ 
& mike\_run & & jenny\_run \\ 
& mike\_sit & & jenny\_sit\\ 
& mike\_fall over & & jenny\_fall over\\ 
& mike\_wave & & jenny\_wave\\ 
& mike\_up & & jenny\_up\\ 
\hline
\multirow{10}{4em}{clothing} & blue hat & \multirow{10}{4em}{large objects} & bee\\ 
& crown & & slide\\ 
& chef hat & & sand\\ 
& pirate hat & & grill \\ 
& sweater hat & & swing \\ 
& silly hat & & tent\\ 
& wizzard hat & & bench\\ 
& horn hat & & christmas tree \\ 
& glasses & & tree\\ 
& sunglasses & & apple tree \\ 
\hline
\multirow{15}{4em}{toys} & baseball & \multirow{7}{4em}{food} & pie\\ 
& glove & & pizza\\ 
& shovel & & hotdog\\ 
& racket & & ketchup\\ 
& kite & &mustard\\ 
& fire & & burger\\ 
& bucket & & coke \\ \hline
& colorful ball & \multirow{8}{4em}{sky objects} & helicopter\\ 
& basketball & & hotair balloon\\ 
& soccer & & cloud\\ 
& tennis ball & & sun\\ 
& football & & lightening\\ 
& frisbee & &rain\\ 
& baseball poll & &rocket\\ 
& balloon &. &plane\\ 
\hline
\multirow{3}{4em}{animal} & bear & \multirow{3}{4em}{animal} & duck \\ 
& cat & & owl\\ 
& dog & & snake\\ 
\hline
\end{tabular}
\end{center}
\caption{List of objects and their hypernyms}
\end{table*}

\clearpage

\section{Implementation Details}

We pretrained the image captioning models using 2 layers of Transformer Decoder with 4 attention heads each, and 512 in internal dimension for 100 epochs each. The dropout rate was 0.5, learning rate started at $1e^{-4}$, on a scheduled decline rate of 0.8 for each 20 unimproved epochs.

We also pretrained the literal listeners and the literal speaker with different disparity datasets. All the weights are fixed before being integrated into the interactive learning phase. During disparity learning, each pair of speaker and listener were trained for 150 epochs, with batch size of 128, learning rate starting at $1e^{-3}$, and on decline at the rate of 0.8 per 20 unimproved epoches. Each experiment is repeated 3 times. The mean and standard deviation were reported in figures. Similarly in the word level training, the model was trained for 200 epochs, and with learning rate starting at 2, and on a scheduled decline rate of 0.8 for each 50 unimproved epochs.

For the efficiency comparison experiment, we used the \textit{combined} test dataset for this experiment, trained each component until 50 unimproved epochs, and selected the top performances within the first 30 minutes of each. All models have reached stable performance by then. All experiments done on a single NVIDIA(R) GeForce(R) RTX 2070 SUPER(TM) 8GB GDDR6 and 10th Gen Intel(R) Core(TM) i9-10900K processor.

\clearpage

% \section{Word Distribution Shift in Word Communication}

% When the speaker and the listener are communicating in words instead of sentences, we observed similar word distribution shift among their description choices. 

% For the $L_1^{d1}$: Hypernym listener in Figure \ref{shift_wh}, the pragmatic rational speaker avoids specific object names, and prioritizing the hypernyms. 

% For the $L_1^{d2}$: Limited Vision listener in Figure \ref{shift_wv}, the pragmatic rational speaker avoids the animal related words, and chose other object such as ``tree'' and ``cloud'' to describe.

% \newpage

% \begin{figure}[t]
%      \centering
%      \begin{subfigure}[b]{0.475\textwidth}
%         \centering
%          \includegraphics[width=0.98\textwidth]{figs/shift_hypernym_wd.png}
%          \caption{$L_1^{d1}$: Hypernym}
%          \label{shift_wh}
%      \end{subfigure}
%      \hfill
%      \hfill
%      \begin{subfigure}[b]{0.475\textwidth}
%          \centering
%          \includegraphics[width=0.98\textwidth]{figs/shift_lv_wd.png}
%          \caption{$L_1^{d2}$: Limited Vision}
%          \label{shift_wv}
%      \end{subfigure}
     
%      \caption{Word Distribution Shift, Communication in Words}
%      \label{shift_wd}
% \end{figure}

\end{document}